\documentclass[mnsc,nonblindrev]{informs3}

\OneAndAHalfSpacedXI

\usepackage{natbib}
 \bibpunct[, ]{(}{)}{,}{a}{}{,}%
 \def\bibfont{\small}%

\usepackage{array}
\usepackage{booktabs}
\usepackage{epigraph}

\TheoremsNumberedThrough     
\ECRepeatTheorems

\EquationsNumberedThrough    

\MANUSCRIPTNO{MS-0001-1922.65}

\begin{document}



\RUNTITLE{Coincidental generation and the misappropriation of likeness}

\TITLE{Coincidental Generation}

\ARTICLEAUTHORS{%
\AUTHOR{Jordan W. Suchow}
\AFF{School of Business, Stevens Institute of Technology, Hoboken, NJ 07030 \EMAIL{jws@stevens.edu}} 
\AUTHOR{Necdet Gürkan}
\AFF{School of Business, Stevens Institute of Technology, Hoboken, NJ 07030}
} 

\ABSTRACT{%
Generative A.I. models have emerged as versatile tools across diverse industries, with applications in privacy-preserving data sharing, computational art, personalization of products and services, and immersive entertainment. Here, we introduce a new privacy concern in the adoption and use of generative A.I. models: that of \textit{coincidental generation}, where a generative model’s output is similar enough to an existing entity, beyond those represented in the dataset used to train the model, to be mistaken for it. Consider, for example, synthetic portrait generators, which are today deployed in commercial applications such as virtual modeling agencies and synthetic stock photography. Due to the low intrinsic dimensionality of human face perception, every synthetically generated face will coincidentally resemble an actual person. Such examples of coincidental generation all but guarantee the misappropriation of likeness and expose organizations that use generative A.I. to legal and regulatory risk.
}%



\maketitle

%


\epigraph{``For years there has been a feeling that the law must afford some remedy for the unauthorized circulation of portraits of private persons.'' ---Warren and Brandeis (1890)}

\section{Introduction}

Generative A.I. models have emerged as a versatile tool across diverse industries, with applications in synthetic data generation, computational art, personalization of products and services, and immersive entertainment. Synthetic data generation, in particular, has emerged as a promising alternative or supplement to standard data masking and anonymization procedures \citep{choi2017generating, ni2020conditional, vietri2020new, chen2021synthetic, xiao2020vae, freiman2017data, yelmen2021creating}. By preserving high-level relationships present in raw data while concealing identifying details, synthetic data generation enables privacy-safe data sharing for various purposes \citep{li2019evaluating}, such as face generation \citep{karras2020analyzing, brown2020language}, test data generation \citep{popic2019data}, world simulations for accelerated learning \citep{ha2018world}, and automated content creation \citep{shu2020fact}. Computational art, another key application area, primarily focuses on generating images for artistic works. Meanwhile, generative text models have become the foundation for numerous natural language processing (NLP) tasks, such as chatbots, text classification systems, and sentiment analysis.

However, the training and use of these models inherently give rise to privacy concerns, potentially jeopardizing the sensitive information of individuals included in the training data. Each year, businesses, governments, research labs, and other institutions release an increasingly large amount of data about their customers, constituents, and participants, among others, in areas ranging from social media to transportation to public health \citep{FederalGovernment, guo2016ms, idrees2017thumos}. The public release of datasets has been an important tool for increasing government transparency \citep{FederalGovernment}, building new technologies \citep{guo2016ms, idrees2017thumos}, enabling replication of research \citep{foster2017open}, and training newcomers in the data sciences \citep{kaggle}. At the same time, the raw data from which publicly released datasets are derived often include considerable detail about people's private lives, such that their direct release would be a threat to the privacy interests of those individuals. In practice, many cases require balancing an important societal need to share and use collected data with the protection of individual privacy \citep{abay2018privacy, vogel2011understanding}.

Historically, machine learning privacy concerns have focused on protecting individual data points within training datasets by using methods like anonymization and differential privacy to mitigate potential risks. Historically, machine learning privacy concerns have focused on protecting individual data points within training datasets. Various methods, including anonymization and differential privacy, have been employed to mitigate potential risks. Three main approaches to privacy preservation can be identified: regulatory measures, access control, and technological solutions. Regulatory measures, such as the European Union General Data Protection Regulation (GDPR) and the California Consumer Privacy Act (CCPA), enhance privacy rights by providing a cause of action against the unauthorized use of personal data to train machine learning algorithms \citep{voigt2017eu, bukaty2019california}. Access control, as exemplified by the National Center for Health Statistics' Research Data Center (RDC), grants restricted-use data access to researchers who justify their need for the restricted data and describe measures to protect the confidentiality of individuals \citep{rdc}. The third approach is technological, with differential privacy (DP) serving as the gold standard of privacy-preserving data release \citep{dwork2014algorithmic}. DP determines the amount of information revealed about entries in the raw data when a query is made to the released dataset \citep{acs2018differentially}. Companies, institutions, and government agencies employ DP for privacy-preserving data practices, including Uber Flex \citep{10.1145/3219819.3220106}, LinkedIn Pinot \citep{rogers2020linkedin}, and US Census Bureau statistics. Neural network models are widely used for differentially private synthetic generation of unstructured data \citep{acs2018differentially, torfi2020differentially, abadi2016deep, ziller2021medical}.

Nevertheless, privacy interests encompass more than just those individuals within the training data, as people outside the dataset may still be affected when a generative model's output unintentionally mimics an existing entity not present in the training data. In particular, we propose that there is a distinct class of individuals with privacy interests in synthetic datasets: those who, despite not appearing in the training dataset, coincidentally resemble records in the synthetic dataset closely enough for the records to be attributed to them.

This paper defines the concept of coincidental generation and investigates its potential to inadvertently misappropriate real-world entities' likenesses.
We focus on synthetic portrait generators as a prime example of this phenomenon and underscore the challenges arising from the low intrinsic dimensionality of human face perception. Subsequently, we scrutinize the legal and regulatory risks confronting organizations that employ generative A.I. and stress the need for heightened awareness and more robust privacy-preserving solutions.

In the ensuing sections, we delve into the technical underpinnings of coincidental generation, assess real-world instances, discuss potential countermeasures, and propose recommendations for tackling the privacy quandaries presented by generative A.I.

\section{Coincidental Generation}
Our foray into coincidental generation begins by defining it. The following definitions presuppose the existence of a generative model that synthesizes artifacts, having been trained on a data set (the ``training set'') of similar artifacts. \textit{Coincidental generation} occurs when a generative model's output is similar enough to an existing entity, beyond those represented in the dataset used to train the model, to be mistaken for it. The following conditions must therefore be met:
\begin{enumerate}
\item \textit{Generated}. The artifact must be synthesized or otherwise created anew by a generative model. An artifact is not generated when the artifact is selected from a set of artifacts previously created, nor when it is a modification of another artifact.
\item \textit{Confusable}. The artifact must be similar enough to an existing entity to produce a likelihood of confusion.
\item \textit{Out-of-sample}. The entity to which the artifact is similar must not be represented in the dataset used to train the generative model.
\item \textit{Coincidental}. The likelihood of confusion must be an unintended side effect of the system's design or training.
\end{enumerate}

\subsection{Establishing the legal and regulatory risks of coincidental generation}

The use of generative AI systems can inadvertently introduce potential legal and regulatory challenges, as users may unintentionally create artifacts that resemble existing intellectual property or real entities. These risks fall into several categories: infringement of registered trademarks, misappropriation of likeness, copyright violations, and unauthorized use of cultural icons or symbols. Users of generative AI models must be aware of these risks and address them responsibly and ethically across various industries.

Users employing AI tools for creative purposes may unintentionally generate content that compromises privacy and publicity rights. For instance, a radio producer who uses a voice-synthesis tool for an advertisement spot might accidentally replicate a famous person's voice. Similarly, a marketer using a face-synthesis tool for ad creative may unintentionally feature a living person's likeness. Medical researchers generating synthetic data without participant PII might inadvertently share PII of a non-participant. Artists creating abstract portraits with AI tools could accidentally generate images resembling celebrities or public figures.

In the realm of intellectual property, users of AI tools can unknowingly create content that violates existing copyrights or trademarks. For example, a graphic designer using a logo-creation tool may unintentionally infringe on a registered trademark, while a content creator employing a text-generation tool might accidentally plagiarize an existing work. Game developers using AI-generated character tools could create characters resembling fictional characters from other games or media, and music composers working with AI-assisted music generators might inadvertently compose melodies resembling copyrighted songs.

In summary, practitioners and researchers using generative AI systems must remain vigilant about the potential legal and regulatory implications of their creations. By addressing the array of potential risks arising from coincidental generation, they can avoid unintended consequences and ensure responsible, ethical use of AI across industries.

\subsection{Formalizing the problem of coincidental generation}
To estimate the discriminability between a synthetic artifact and its nearest neighbor among all past, current, or future entities, we rely on the statistical relationship between sample size, the dimensionality of the feature space, and the distribution of $n$-th nearest neighbor distances in the feature space.

Before we continue, note that the feature space considered here is not the space defined by pixel intensities or the space defined by the latent variables of the synthetic artifact generator. Rather, it is the human perceptual feature space—one in which points are instances, distributions over points are identities, and distances between points give the similarity of the instances in perceptual units, e.g., just noticeable differences (JNDs) in identity. Though these three metric spaces are not unrelated, they are not equivalent, and one must be careful to distinguish between them in the analysis. Most relevant to the privacy concerns we raise here are the perceptual similarity (or discriminability) between synthetic artifacts and real entities.

Assume that artifacts are uniformly distributed in human perceptual feature space, scaled to unit volume (see section "Setting $c$" for a discussion on relaxing this assumption to accommodate perceptual expertise and other effects of perceptual learning). Nearest neighbor distances of points distributed uniformly in a unit volume depend both on the number of sampled points and on the dimensionality of the space. Bhattacharyya \& Chakrabarti (2008) showed that the mean $n$-th nearest neighbor distance among $N$ points distributed uniformly in a unit volume is given by
\begin{equation}
\langle r_n (N)\rangle = \frac{[\Gamma(\frac{D}{2})+1)]^{\frac{1}{D}}}{\pi^{\frac{1}{2}}} \frac{\Gamma(n+\frac{1}{D}))}{\Gamma(n)} \frac{\Gamma(N)}{\Gamma(N+\frac{1}{D})},
\end{equation}
where $D$ is the dimensionality and $\Gamma$ is the gamma function \citep{bhattacharyya2008mean}. When $N \gg n$, this value can be approximated by
\begin{equation}
\langle r_n (N)\rangle\approx (\frac{n}{N})^{1/D}.
\end{equation}

In addition to the dimensionality and the number of points distributed in the space, the discriminability between appearances also depends on the viewer's sensitivity \citep{green1966signal}. In particular, when the viewer is highly sensitive, two points that are close together will nonetheless be reliably distinguished when they come from two identities. In contrast, when the viewer is insensitive, two points that are far apart may nonetheless be indistinguishable to the viewer. To capture this notion of viewer sensitivity, we scale the unit volume by the constant $d^\prime$, a measure of perceptual discriminability from signal detection theory, converting the unit volume assumed in Equation 2 to units of perceptual discriminability. Because $d^\prime$ is defined at the level of the observer, but we wish to make a claim about a generic reasonable observer, we consider $\bar{d}^\prime$, the average discriminability across observers.

Ultimately, to determine whether synthetic artifact generators preserve the privacy of all individuals, we ask whether
\begin{equation}
\bar{d}^\prime \langle r_1 (N_\forall)\rangle < c,
\end{equation}
where $N_{\forall}$ is the number of all entities and $c$ is the threshold perceptual distance within which the synthetic artifact sufficiently resembles an actual entity that fails to preserve its privacy

Defining a statistical model of $n$-th nearest neighbor discriminability converts an otherwise nebulous problem into several less-nebulous estimation problems. We proceed by estimating each of the parameters in turn for the case of synthetic face generation.

\section{A case study on synthetic face generation}
Consider embarking on a thought experiment designed to reveal the nature of face space's dimensionality, and in doing so, engage with the intuitive understanding of human facial similarity. This thought experiment relies on the mathematical properties of nearest neighbor distances in low- versus high-dimensional spaces. Specifically, it highlights the fact that in low-dimensional spaces, nearest neighbors are substantially closer than random neighbors, whereas in high-dimensional spaces, nearest neighbors are only marginally closer than random neighbors.

Envision a friend or family member whose visage is intimately familiar to you, their facial features ingrained in your memory. With their countenance in mind, consider the following question: have you ever encountered a stranger bearing even a passing resemblance to this person, someone you might momentarily confuse with your acquaintance from a distance?

In a world governed by high-dimensional facial representations, the likelihood of such an occurrence would be vanishingly small. Not only would it be improbable to encounter a stranger with even a hint of resemblance to your friend or family member, but distinguishing varying degrees of similarity between strangers would be virtually impossible.

Yet, the reality we experience is markedly different. Ours is a world teeming with doppelgängers, twin strangers, celebrity lookalikes, and cases of mistaken identity. These phenomena, pervasive in our daily lives, underscore the fact that human face space is indeed of low dimensionality, with a rich tapestry of resemblances woven throughout our social fabric.

By engaging with this thought experiment, one can appreciate the inherent complexity of human facial similarity and the intriguing nature of the low-dimensional space that governs our perception of faces. This understanding not only fosters a deeper appreciation for the subtleties of human recognition but also highlights the challenges and opportunities that lie ahead in the realm of machine learning and facial recognition technologies.

\subsection{A statistical model of face generation}
In this section, we delve into a case study on coincidendental generation in synthetic face data. The core argument is that the manifold of actual human appearances is so densely packed that the nearest neighbor to any randomly chosen point in this space, i.e., the individual whose appearance most closely resembles a synthetic portrait generated by a well-trained model, is likely to be virtually indistinguishable from the synthetic portrait itself. By presenting a statistical analysis of the variation in human appearance, we argue that every synthetic portrait inevitably captures the likeness of an actual individual, be it someone from the present, past, or future. This case study serves to illuminate the privacy ramifications arising from the coincidental generation of synthetic data and highlights the necessity of reevaluating privacy preservation strategies in the context of synthetic data generation.

Modern machine-learning solutions to the problems of face detection, localization, verification, recognition, and related tasks train deep neural networks on large datasets of photographs of people’s faces. The photos in these image sets depict individuals with privacy interests that may diverge from the interests of those who curated the dataset, trained the algorithms, or deployed the resultant technologies. One possible method to mitigate these privacy concerns is to generate synthetic data to train models. For example, in the case of faces, it is often possible to adapt a variational autoencoder (VAE) \citep{yan2016attribute2image, hou2017deep} or a generative adversarial network (GAN) \citep{karras2019style,karras2020analyzing} for use as a generator of synthetic portraits, essentially laundering (or otherwise obscuring) the identities of those individuals depicted in the photos used to train the generator. It has been argued that synthetic portraits may provide a solution to protecting the privacy of individuals with respect to the training of machine-learning models by obviating the need for the use of new photographs of people in the training of face-recognition technologies, broadly construed.

At the same time, solutions to other problems, such as face anonymization through face replacement \citep{sun2018hybrid}, seek to anonymize a portrait by replacing the depicted person's visage with that of a synthesized identity. Though most major social media platforms have policies in place to safeguard against misuse of other people's photographs and enable users to report cases of impersonation if their photographs are being used without consent \citep{twitter, facebook}, synthetic portrait generators have been used to infiltrate the intelligence community on LinkedIn, to publish right-wing propagandist articles on news platforms, and to harass people on the internet \citep{nytimes}. 




Notions of privacy such as $k$-anonymity \citep{samarati1998protecting} and differential privacy \citep{abadi2016deep, torkzadehmahani2019dp} seek to preserve or characterize the privacy of those individuals depicted in the training set. In contrast, we focus here on a different set of individuals with privacy interests in the images: individuals who are not depicted in the training set but who coincidentally resemble synthetic portraits generated by the trained model. The crux of the argument is that the collection of actual appearances is so densely arranged in face space that the nearest neighbor to a randomly selected point in the space (i.e., the person who most closely resembles a synthetic portrait sampled from a well-trained model) is likely to be indistinguishable from them. In particular, through a statistical argument about variation in human appearance, we argue that every synthetic portrait captures the likeness of an actual current, past, or future person. 

\section{Estimating $N_{\forall}$}
Here, we estimate $N_{\forall}$, the total number of people who may be the target of coincidental generation.

Is $N_{\forall}$ equal to the number of identities in the training set? No. Even the largest image sets used to train modern synthetic face generators contain orders of magnitude fewer faces than are necessary to make the universal argument we consider here. Image sets used to train modern synthetic portrait generators contain only a small fraction of current, past, and future identities. For example, Flickr-Faces-HQ (FFHQ), the image set used to train StyleGAN2, a popular synthetic portrait generator \citep{karras2020analyzing}, contains 70,000 images at therefore at most 70,000 unique identities. The CelebA image set contains 10,177 identities. Labeled Faces in the Wild also contains on the order of 10,000 unique identities.

Global demographic estimates put the number of currently living humans at 7.8 billion. Scientists have estimated that there have been roughly 100 billion humans who have ever lived \citep{kaneda2018many}. Extrapolating from these figures to estimate the number of people who will ever live is a more challenging task. Under a statistical formulation of the Copernican principle, whereby we are not privileged observers of the universe, and assuming an uninformative prior over the duration of humanity \citep{gott1993implications, bostrom2013anthropic}, one can estimate that there will eventually be 200 billion humans who have ever lived. These figures are one-hundred-thousandfold, 1.4-million fold greater, and 2.9 million fold greater than the 70,000 faces in FFHQ, respectively.

In contrast to these staggering estimates of the number of the extant number  of people, the number of identities known to the average person is quite small. A limit to precision can be achieved as with abundance estimates in other domains such as the number of habitable planets \citep{lissauer1999common} and number of species on Earth \citep{mora2011many}. \citeauthor{jenkins2018many} (\citeyear{jenkins2018many}) combine separate measures of recall and recognition to show that people know around 5000 faces on average and that individual differences are large. 

The average number of identities that a person encounters in a lifetime is necessarily at least that large, but estimates of the quantity are hard to come by and likely vary across individuals by several orders of magnitude. Thus the average person recognizes as familiar only 1 in roughly 7.8 million faces and has seen perhaps 1 in roughly 1 million. Intuitions about similarity developed at such miniscule scales are unlikely to hold at the full scale of humanity.

\section{Estimating $D$}

Trained generative models of human identity and appearance are typically high dimensional \citep{gulrajani2016pixelvae, karras2020analyzing}, with representations such as those learned by the StyleGAN2 architecture representing each portrait as a point in a space with hundreds to thousands of dimensions \citep{karras2020analyzing}. In contrast, human representations of faces have low dimensionality, with empirical estimates implying a dimensionality between 7 and 12 dimensions \citep{ryali2020leveraging, pincus2020estimating}. Note that, or any given person, there are many unrelated people who look similar to them. Consider, for example, the phenomena of ``twin strangers'' and ``doppelgängers'', where strangers look almost impossibly alike. Next, we note that the very concept of strong resemblance exists only in low dimensional spaces. When points are well-distributed in high dimensional spaces, nearest neighbors (i.e., a person and their closest doppelgänger) are nearly as far apart as randomly selected pairs of points (i.e., a person and a random stranger). By contradiction, face space must be low-dimensional. 

How does the dimensionality affect nearest neighbor distances? At one extreme, where the space is one-dimensional, an increase in the number of points leads to an equivalent decrease in the average distance to the nearest neighbor. At the other extreme, however, where the space is high dimensional, an increase in the number of points has a negligible effect on the average distance the nearest neighbor.

\section{Setting $c$}

Portrait artists often attempt in their artwork to capture someone’s \textit{likeness}, the aspects of a person’s appearance that support recognition. Importantly, the goodness of a likeness is subjective, depending not on any directly measurable feature of the portrait, but on the impression that it leaves on the viewer. For example, a photorealistic portrait identical to a target’s image in every way except one may nonetheless fail to produce a good likeness. In contrast, an abstract portrait (e.g., in the style of Stanley Chow) or a caricature (e.g., in the style of Al Hirschfeld) may vary from a target’s image in almost every way yet produce a likeness better than any photograph ever could.

A portrait may also leave different impressions in different viewers because of their varying visual expertise with respect to an individual’s appearance. For example, a portrait may represent a good likeness of a person only to a viewer with moderate expertise in recognizing that person, at the same time being rejected as a likeness by those more familiar with the person’s appearance. 

A good likeness can therefore be reasonably defined in several ways. At one extreme, a portrait could be said to represent a good likeness of an individual if any viewer would accept it as such. At the other extreme, a portrait could be said to represent a good likeness of an individual only if the viewer most intimately familiar with the person’s appearance would accept it as such. In between the extremes are a gamut of likeness standards. 

Which standard of likeness is relevant in a given setting will depend on the privacy implications. In a setting where preserving the privacy of individuals is critical, it may be important that no viewer mistakenly accept the purportedly synthetic portrait as a likeness of the individual. In settings where privacy interests are less important, it may be enough to demonstrate that a typical viewer would not reasonably accept it as such.

Therefore, when we ask whether a synthetic photo protects a certain individual’s privacy interests, we will consider not whether the photo depicts the person (in the sense of being a visually conveyed reference to them), which it deliberately does not, but whether the photo contains a likeness of the individual from the perspective of the relevant viewer.

\subsection{False alarms by pretrained machine vision systems}

We input samples from the synthetic portrait generator to facial recognition systems pretrained on images of celebrities and other famous or prominent people (e.g., Amazon Rekognition). Each input image generates an output that is either a non-match decision or a positive match with an associated measure of confidence. Because none of the photos depict real persons, all matches found by the face-recognition system are false alarms. How many samples are close enough to be incorrectly identified as a celebrity, and with what confidence? 

In a first analysis, we analyzed these false alarms, finding that the face-recognition system falsely recognized 144/1000, or $14\pm2.2 \%$ (mean $\pm$ \textsc{SE}) of the synthetic portraits. Fig. 1 plots the distribution over confidence levels associated with these false alarms. (N.B.: the face-recognition system returns a positive match only when the confidence is greater than 0.50). Fig. 1 plots the false alarm rate as a function of the criterion level of confidence.

In a second analysis, we curated a paired image set that includes one actual photo of each of the identities reported in false alarms. We then measured the face-recognition system's confidence in identifying the celebrities in the combined image set, producing a label and confidence level associated with each photo. We then performed an analysis that simulates a discrimination task where the face-recognition system is asked to determine which photo --- celebrity or synthetic false alarm --- is the actual celebrity. To accomplish this, we computed the proportion of synthetic--celebrity photo pairs for which the internal criterion is higher for the celebrity than for the synthetic photo. We found that the system correctly identified the celebrity in $85\%$ of image pairs. Combining the two analyses, then, we find that of the $14\%$ of random samples from the synthetic portrait generator that were falsely identified as a celebrity, $15\%$ were more readily accepted as depictions of those celebrities than actual photographs, implying that $2.1\%$ of synthetic portraits are within 1 JND of a celebrity from the perspective of a well-trained machine vision system. Given that Amazon Rekognition recognizes approximately 8M celebrities, this represents only one ten thousandth of the total population.

\begin{figure}
  \label{fig:rekognition}
  \centering
  \includegraphics[width=5cm, height=5cm]{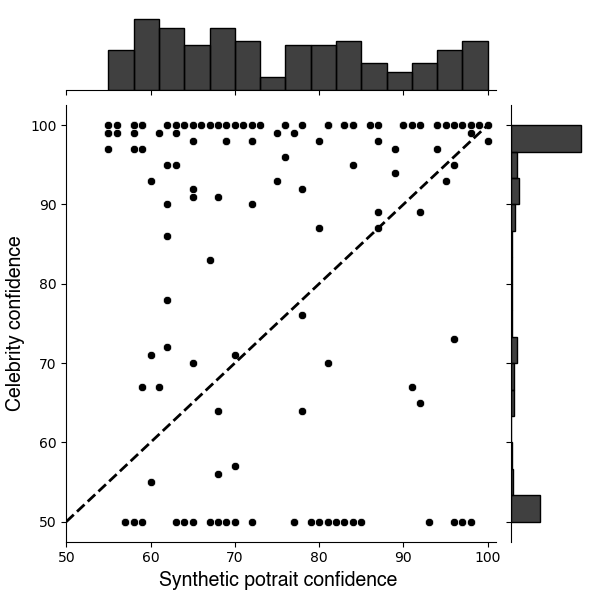}
  \caption{Commercial face-recognition systems falsely recognize synthetic portraits as actual people.}
\end{figure}

\section{Discussion}

For approaches to mitigating the risk of coincidental generation, one might look to other area of creative design where overlap between new and existing artifacts is of concern. One such area is logo design. In this context, the possibility of coincidental resemblance to existing logos poses a risk of unintentional copyright or trademark violations, not one of misappropriation of likeness. The intrinsic dimensionality of logo design, while arguably higher than that of human faces (e.g., logos consisting of a realistic face is only a small subset of the logo design space), remains limited due to the need for logos to be simple, memorable, and easily recognizable. This constraint increases the likelihood of coincidental generation. To mitigate this risk, trademark search is a standard practice in the branding process, helping to identify potential conflicts and ensure the originality of a new logo. And indeed, generative A.I. tools for logo design routinely include databases of tens of thousands of previously designed logos, which serve both as a source of inspiration and fodder for the model's generative designs, but also a check on coincidental generation through the explicit inclusion of a step in the algorithm that computes a visual similarity score between proposed designs and each logo in the database. No such database is possible in the case of faces, for both practical reasons and the more fundamental ethical issue inherent in creating large-scale databases of identities.

\subsection{Relation to hash collisions}
Hash functions maps between input data and an output label of a fixed length, and are commonly used in data management and cryptography to uniquely label the input data without needing to handle it directly (e.g., to prevent disclosure of personally identifiable information or to avoid needing to directly process large files). A hash collision arises when two distinct inputs produce the same hash value under a given hashing function. Coincidental generation and hash collisions thus analogous: both involve unintended similarity of the output to an existing entity.

Of considerable importance in the design and analysis of hashing algorithms is the probability of a hash collision \citep{peyravian1998probabilities}. Dimensionality and the number of values per dimension play a crucial role in determining the probability of hash collisions. For example, MD5 and SHA hashing algorithms produce fixed-length output hash values, with MD5 generating 128-bit hashes and SHA-1 producing 160-bit hashes. The number of possible hashes, and thus the probability of a hash collision, depends on both their length and the number of symbols used (e.g., 16 hexadecimal symbols for MD5 and SHA). As the number of possible hash values increases, the probability of collisions decreases. However, no hashing algorithm can be completely collision-free, as the pigeonhole principle guarantees collisions when there are more input values than unique output hash values.

\subsection{Relation to the Infinite Monkey Theorem}

The infinite monkey theorem posits that a monkey hitting keys at random on a typewriter for an infinite amount of time will almost surely produce any given text, such as the complete works of Shakespeare. This theorem highlights the concept of coincidental generation in the context of text, where the sheer volume of random input combinations, given infinite time, can eventually result in the generation of recognizable content.

However, there are important differences between text and faces when it comes to coincidental generation. In the case of text, the probability of generating a specific meaningful sequence is low, as the number of possible character combinations is vast. Consequently, the infinite monkey theorem relies on the assumption of infinite time to guarantee the generation of any desired text.

In contrast, human faces have a lower intrinsic dimensionality, making the space of possible faces much smaller than the space of possible text combinations. Consequently, the probability of generating a face that closely resembles an existing person is relatively higher. Furthermore, generative AI models are trained to produce realistic faces, which further increases the probability of coincidental generation.

\section{Conclusion}
We propose that there is a distinct class of individuals with privacy interests in synthetic datasets: those who, despite not appearing in the training dataset, coincidentally resemble records in the synthetic dataset closely enough for the records to be attributed to them. Using synthetic portrait generators as a case study, we argue that the collection of actual appearances is so densely arranged in face space that every synthetic portrait will necessarily co-incidentally capture the likeness of at least one actual current, past, or future person. In contexts such as facial anonymization where the goal is to mask the identity of a person, use of synthetic portraits is ill-advised because they are likely to implicate the presence of other persons. We consider the privacy implications of coincidental generation for the use of synthetic data generators.




\newpage

\bibliographystyle{informs2014} 
\bibliography{references.bib}

\end{document}